\documentclass[conference]{IEEEtran}
\IEEEoverridecommandlockouts
\usepackage{cite}
\usepackage{amsmath,amssymb,amsfonts}
\usepackage{algorithmic}
\usepackage{graphicx}
\usepackage{textcomp}
\usepackage{makecell}

\usepackage{pifont}
\usepackage{xcolor}
\usepackage{xurl}
\usepackage{subcaption}
\usepackage[export]{adjustbox}
\usepackage[bookmarks=true,breaklinks=true,letterpaper=true,colorlinks,linkcolor=blue,citecolor=blue,urlcolor=cyan]{hyperref}

\usepackage{url}
\usepackage{xspace}        
\usepackage[a4paper, total={184mm,239mm}]{geometry}
\def\BibTeX{{\rm B\kern-.05em{\sc i\kern-.025em b}\kern-.08em
    T\kern-.1667em\lower.7ex\hbox{E}\kern-.125emX}}
\usepackage{authblk}

\author[$\dagger$]{Yu-Shun~Hsiao$^{*}$}
\author[$\dagger,\ddagger$]{Zishen~Wan$^{*}$\thanks{$^{*}$These two authors contributed equally, listed in alphabetical order.} }
\author[$\dagger,\P$]{Tianyu~Jia\thanks{$^{\P}$This work was done while the author was at Harvard University.}}
\author[$\dagger$]{Radhika~Ghosal}
\author[$\dagger$]{Abdulrahman~Mahmoud}
\author[$\ddagger$]{\\Arijit~Raychowdhury}
\author[$\dagger$]{David~Brooks} 
\author[$\dagger$]{Gu-Yeon Wei}
\author[$\dagger$]{Vijay~Janapa~Reddi \vspace{-0.05in}}

\affil[$ $]{$^{\dagger}$Harvard University \hspace{0.05in} $^{\ddagger}$Georgia Institute of Technology \hspace{0.05in} $^{\P}$Peking University\vspace{-0.2in}}

\begin{document}

\title{MAVFI: Transient Fault Analysis Framework for Micro Aerial Vehicle Reliability} 
\title{Improving the Safety and Resilience of Autonomous UAVs}
\title{MAVFI: An End-to-End Framework for Systematically Characterizing and Analyzing the Resilience of UAVs to Hardware Faults} 
\title{Robot Operating System (ROS) Fault Injection\\with Anomaly Detection and Recovery}
\title{Silent Data Corruption in Robot Operating System (ROS):\\A Case for End-to-End System-Level Fault Analysis Using Autonomous UAVs}
\title{Improving the Safety and Resilience \\of Autonomous UAVs}
\title{MAVFI: An End-to-End Fault Analysis Framework with Anomaly Detection and Recovery \\for Micro Aerial Vehicles \vspace{-0.05in}}
\maketitle

\begin{abstract}
Safety and resilience are critical for autonomous unmanned aerial vehicles (UAVs). We introduce MAVFI, the micro aerial vehicles (MAVs) resilience analysis methodology to assess the effect of silent data corruption (SDC) on UAVs' mission metrics, such as flight time and success rate, for accurately measuring system resilience. To enhance the safety and resilience of robot systems bound by size, weight, and power (SWaP), we offer two low-overhead anomaly-based SDC detection and recovery algorithms based on Gaussian statistical models and autoencoder neural networks. Our anomaly error protection techniques are validated in numerous simulated environments. We demonstrate that the autoencoder-based technique can recover up to all failure cases in our studied scenarios with a computational overhead of no more than 0.0062\%. Our application-aware resilience analysis framework, MAVFI, can be utilized to comprehensively test the resilience of other Robot Operating System (ROS)-based applications and is publicly available at~\url{https://github.com/harvard-edge/MAVBench/tree/mavfi}.


\end{abstract}


\section{Introduction}
\label{sec:intro}

Silent data corruptions (SDCs) have become an important problem~\cite{li2017understanding}. It has been shown as a major issue for server scale systems~\cite{dixit2021silent,hochschild2021cores}. However, there are many emerging application areas where SDC effects extend beyond just computational reliability into safety. Such an emerging area is unmanned aerial vehicles (UAVs) where resilience \textit{and} safety are critical. 

\begin{figure}[t!]
        \centering
        \includegraphics[width=\columnwidth]{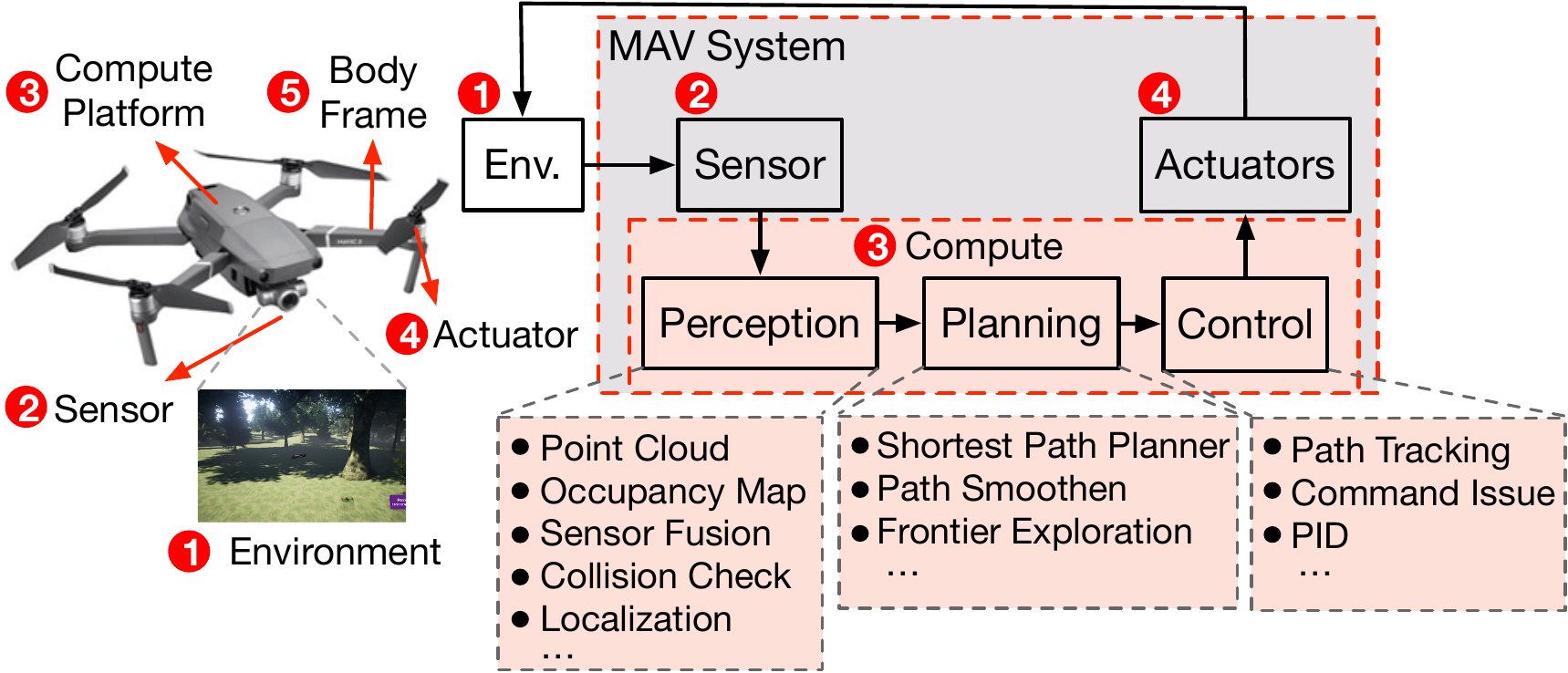}
        \vspace{-0.2in}
        \caption{
        End-to-end perception-planning-control (PPC) computing paradigm. Each PPC stage contains multiple kernels, and we study the safety and resilience of the end-to-end pipeline.}
        \label{fig:PPC_pipeline}
        \vspace{-15pt}
\end{figure}

SDCs caused by external radiation and voltage noise~\cite{mukherjee2005soft} in the computational element like the compute subsystem present a major threat to the safe deployment of UAVs~\cite{ingenuity}, whose deployment can be impeded in many real-world scenarios. To assess SDC's impact on UAVs, \textbf{we propose the first system-level metrics for fault characterizations on a ROS-based autonomous system}. The autonomous UAV consists of an {end-to-end} perception-planning-control (PPC) pipeline (Fig.~\ref{fig:PPC_pipeline}) that generates real-time flight commands based on the environment. The PPC pipeline is the decision-making center for a UAV to maneuver safely. A SDC could cause a UAV to detour or crash. 
%
%

Prior works adopt redundancy~\cite{talpes2020compute} at the hardware or software level to improve autonomous vehicles (AVs) resilience. 
While existing techniques are effective, they are infeasible for size, weight, and power (SWaP)-constrained AVs such as UAVs due to both the power and form factor limitations of a UAV system. Recent software technique~\cite{hari2021making,chen2021low} for the resilience of deep neural networks (DNNs) on GPU does not apply to UAVs that typically do not have access to power-hungry GPUs onboard. Besides, UAVs have a strict limit on total flight time due to the limited onboard battery capacity. Therefore, UAVs need a lightweight fault mitigation technique to prevent harmful SDC from detouring or even crashing the UAV without degrading the total flight time and system availability.

\textbf{We propose two anomaly detection and recovery methods.} First, we propose a Gaussian-based anomaly detection (GAD) and recovery mechanism 
(\S\ref{subsec: gaussian detection}). We leverage the characteristics that UAVs' movements are continuous and each temporal transition of inter-kernel states is close to a Gaussian distribution. This technique features Gaussian-based range detectors for each inter-kernel state that cease error prorogation once an outlier is detected. Second, we propose and evaluate an autoencoder-based anomaly detection (AAD) technique to improve UAV 
resilience (\S\ref{subsec:autoencoder}). 
AAD adopts a neural network-based autoencoder to learn normal UAVs' kinematics and detect anomalies according to the reconstruction error of the input delta values, leveraging correlation among inter-kernel states. 

We evaluate the effectiveness of the two techniques across four 
vastly different types of environments on two compute platforms with the simulated micro aerial vehicle (MAV)~\cite{boroujerdian2018mavbench}. 
Our results demonstrate that the Gaussian-based technique recovers up to 89.6\% of failure cases, 
and \textbf{the autoencoder-based method can recover 100\% failures} in the best-case scenario. Moreover, the overhead of AAD is only up to 0.0062\% and much smaller than 2.22\% of the Gaussian-based technique. 

We also show that \textbf{our autoencoder-based anomaly detection and recovery technique can reduce UAV flight-time energy usage by up to 1.91$\times$ more than traditional redundancy-based hardware solutions (e.g., DMR, TMR)}. The redundancy-based solutions increase the weight and form factor of UAVs and lead to performance overheads. Regarding quality-of-flight (QoF) efficiency metrics, 
the Gaussian-based technique can recover the SDC-degraded flight time by up to 63.5\% and 73.0\% for the 
autoencoder-based technique. 

In summary, the contributions of this work are as follows:
\begin{itemize}
    \item We present MAVFI, the first ROS-based application-aware resilience analysis framework, to analyze UAVs' fault tolerance characteristics with proper system-level metrics.
    \item We conduct fault tolerance characterizations of the end-to-end PPC pipeline and show that application-aware metrics are essential to understanding fault's impact on kernels. 
    \item We propose {two low-cost anomaly error detection and recovery schemes} and evaluate them on different UAV configurations, and show that SDC impact on safety can be rectified in real-time with negligible overhead in ROS.
\end{itemize}

\section{MAVFI Fault Injection Framework}
\label{sec:methodology}

To analyze SDCs' impact on UAVs, we first and foremost need a fault injection framework in the ROS middleware for injecting faults into the end-to-end UAV application pipeline to assess their impact systematically. This section presents MAVFI that supports fault injection with QoF metrics for evaluation.

\begin{figure}[t!]
\vspace{-2.25em}
\centering\includegraphics[width=\columnwidth]{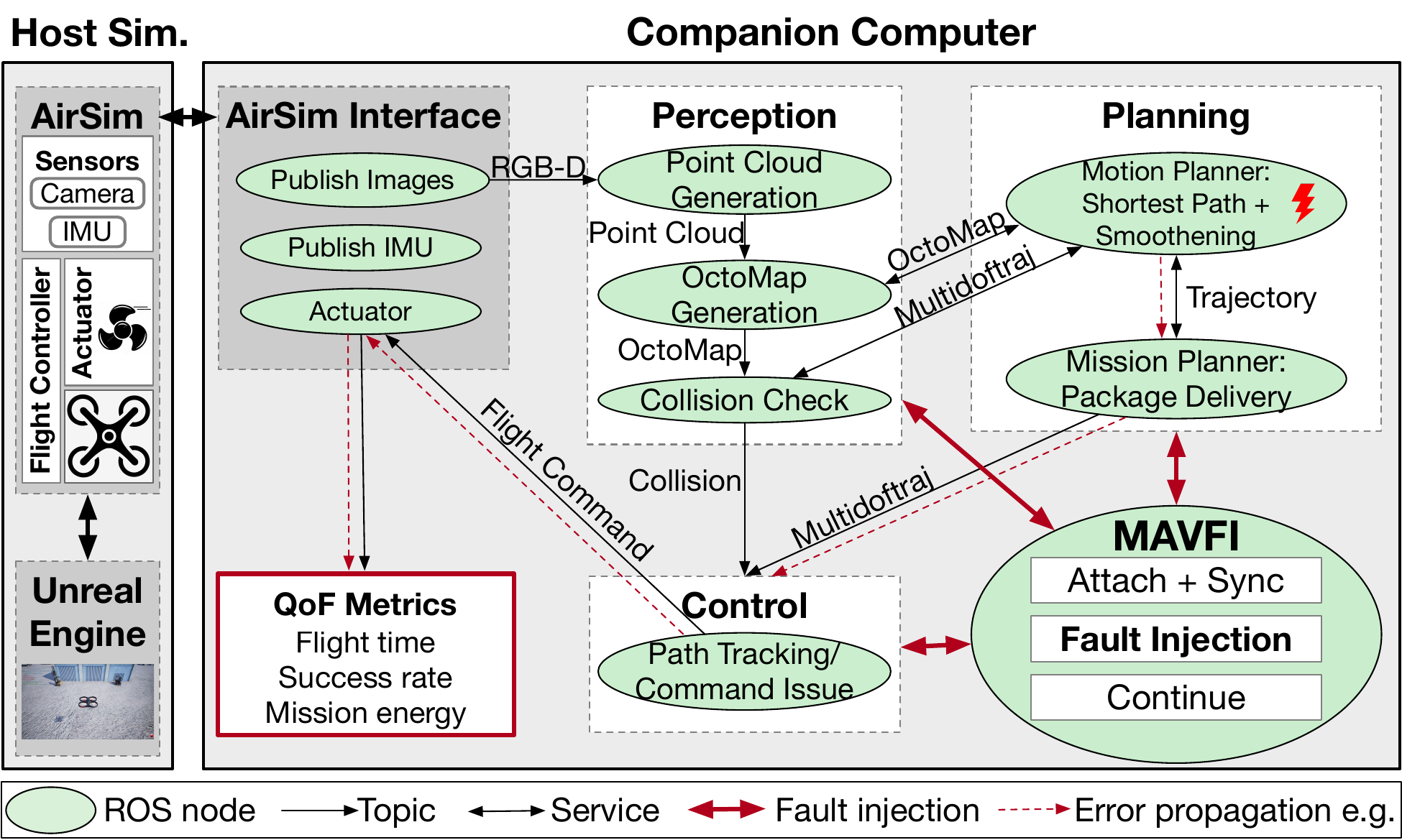}
        \caption{End-to-end MAVFI resilience analysis framework.}
        \label{fig:Inject-overview}
        \vspace{-0.1in}
\end{figure}

\subsection{MAV Fault Injector Implementation}
\label{subsec:MAVFI implementation detail}
Fig.~\ref{fig:Inject-overview} illustrates our  fault injection infrastructure for a ROS-based UAV system. It includes the simulated environment and UAV on the host system. The UAV's PPC pipeline is integrated with MAVFI on a ``companion'' computer. The companion computer processes high-level tasks, while a microcontroller typically handles the low-level flight controller commands. 

Each PPC stage contains one or multiple ROS nodes. Each ROS node comprises a single compute kernel, such as a motion planner. ROS node communicates through ROS topics (one-to-many communication) and ROS services (one-to-one communication). MAVFI is built as a ROS node to maintain our framework's portability, and it leverages the ROS communication protocol and Linux system calls to inject faults. Fig.~\ref{fig:Inject-overview} illustrates an error propagation example when a fault is injected in the \textit{Motion Planner} kernel---it manifests as a corruption of execution in \textit{Multidoftraj}, \textit{Trajectory}, which eventually corrupts a flight command and impacts the quality of flight (QoF).  

To establish UAV experiments, we integrated the fault injection with a ROS-based UAV simulator, MAVBench~\cite{boroujerdian2018mavbench}. MAVBench includes Unreal Engine to simulate the surrounding environment, AirSim to capture a UAV's kinematics, and PPC pipeline to generate flight commands in real-time. The PPC pipeline processes the sensor data and generates flight commands continuously until the mission is complete. Finally, the real-time mission QoF metrics are recorded. Although we use a MAV as an example, the fault analysis methodology is broadly applicable to any ROS-based AV use case.

\subsection{Fault Model}
\label{subsec:MAVFI implementation detail}

MAVFI emulates {instruction-level fault injection}, which is in line with prior work~\cite{wei2014quantifying,jha2019ml,mahmoud2019minotaur}. A limitation is that it does not consider faults in the memory and caches. Typically, ECC is used to protect caches and memory for robots as is the case with the NVIDIA TX2/Xavier series of hardware, which we use. We also assume no faults in the processor's control logic, which constitutes a small portion of the processor~\cite{wei2014quantifying,mahmoud2019minotaur}.

\section{End-to-End PPC Pipeline Fault-Tolerance}
\label{sec:algorithm}
This section presents the fault tolerance analysis with application-aware system-level performance metrics. We explore how errors would impact a single kernel and propagate through the whole PPC pipeline to affect UAV QoF metrics.

\begin{figure}[t!]
\vspace{5pt}
\centering
  \subfloat[Flight time.]{
	\begin{minipage}[c][0.25\columnwidth]{
	   0.5\columnwidth}
	   \centering
	   \includegraphics[width=1.0\textwidth]{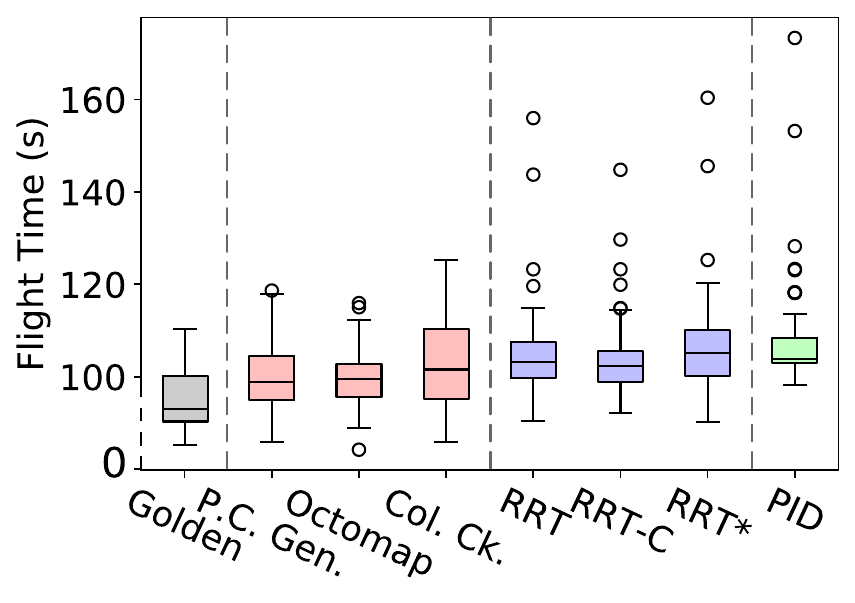}
	   \label{fig:PPC_time}
	\end{minipage}}
  \subfloat[Flight success rate.]{
	\begin{minipage}[c][0.25\columnwidth]{
	   0.5\columnwidth}
	   \centering
	   \includegraphics[width=1.0\textwidth]{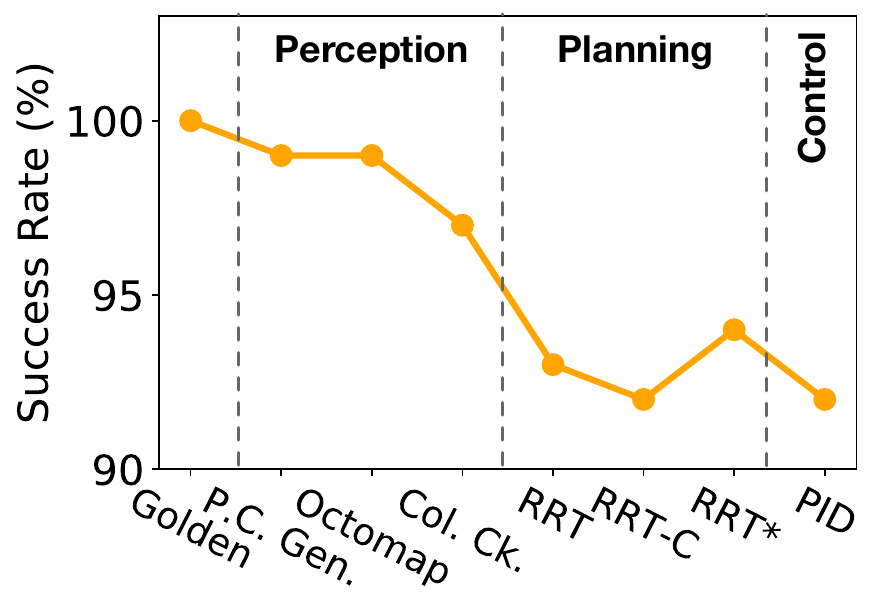}
	   \label{fig:PPC_success}
	\end{minipage}}
	\vspace{-5pt}
\caption{Application-aware and system-level end-to-end fault tolerance analysis with an instruction-level fault injector.} 
\label{fig:MAVFI_fault_characterization}
\vspace{-15pt}
\end{figure}

\subsection{End-to-End, System-level Fault Tolerance Analysis}
\label{subsec:MAVFI_results}

We conduct the \textit{first} end-to-end system-level analysis on how kernel errors would propagate through PPC pipelines and impact UAV performance. We perform 100 fault injection runs per kernel. Besides the fault injections we perform, 100 error-free experiment runs are defined as \textit{Golden}. In each experiment, all kernels in the PPC pipeline are launched by ROS to complete a given navigation task. Only one of the kernels would have a one-time single-bit fault injection during each flight mission for fault injection runs. Without loss of generality, we limit our discussion to a navigation task in the \textit{Sparse} environment here. More results are demonstrated
in~\S\ref{sec: results}.



\textbf{Finding: The visual perception stage is the least critical when a SDC manifests as the downstream perception tasks make up for it.} The typical PPC pipeline includes \textit{Point cloud generation (P.C. Gen.), OctoMap, Collision check (Col. Ck.)} for perception, \textit{RRT*} for planning, and \textit{PID} for control. Two other planning algorithms are evaluated, i.e., \textit{RRT} and \textit{RRTConnect}. 

Prior works often tend to overly focus on error resilience of the perception stage~\cite{li2017understanding,hari2021making}. However, as Fig.~\ref{fig:MAVFI_fault_characterization} shows, for the perception stage both \textit{Point Cloud Generation} and \textit{OctoMap} have little to negligible impact on the system.
The reason that \textit{OctoMap} is resilient in the end-to-end analysis is that even if an occupied voxel is corrupted and mistaken as a free voxel, all other voxels around it are still occupied. So this means that the UAV can still determine obstacles' locations provided the \textit{OctoMap}'s resolution to make the correct flight action decisions. The critical observation is that this is a counter-intuitive observation that is difficult to discover without end-to-end analysis. \textit{Collision Check} is critical in the perception stage since a false alarm can lead to  re-planning or collisions. 

\textbf{Finding: Planning and control are more critical than perception.} The corrupted outputs (e.g., yaw, roll, pitch, velocity) from these \textit{Planning} and \textit{Control} stages can directly lead to a detour or crash of the UAV. From Fig.~\ref{fig:PPC_time}, even though the average flight time is similar, the range of \textit{RRT, RRTConnect, RRT*, and PID} is much wider than \textit{Octomap} and \textit{Golden}. The error propagation of the corrupted execution results could greatly increase the flight time by up to 57.3\% and even lead to degradation of success rate by up to 8\% as shown in Fig.~\ref{fig:PPC_success}. 

Hence, the planning and control stages are more critical than the perception stage from an end-to-end application perspective.


\subsection{Error Propagation Across PPC Stages}
\label{subsec: cross-stage}
\begin{figure}[t!]
        \centering\includegraphics[width=\columnwidth]{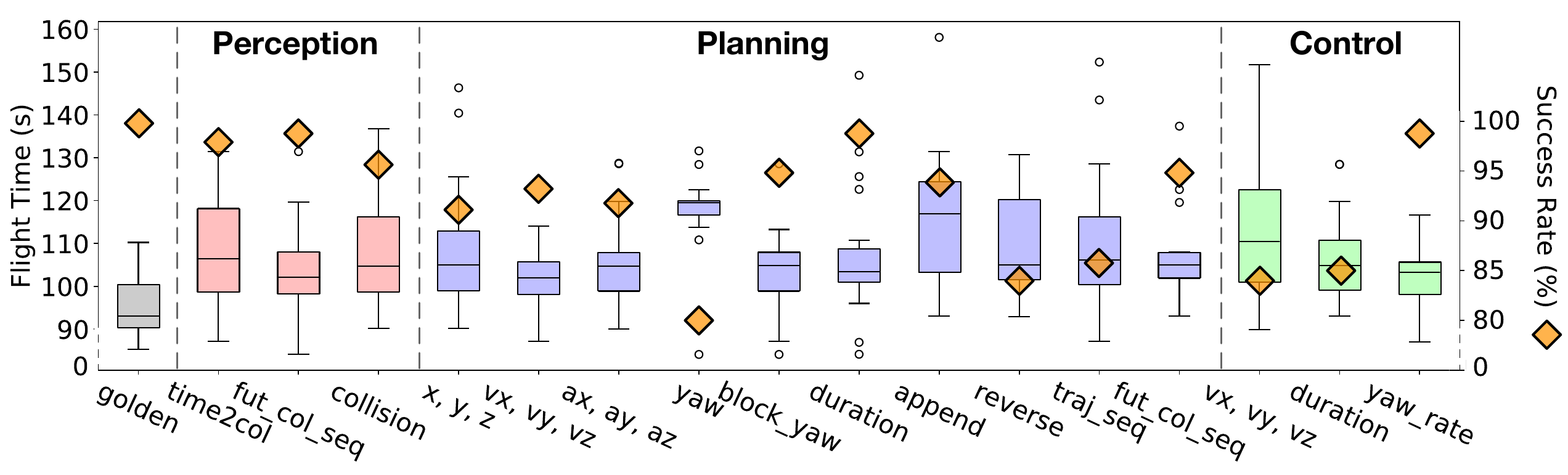}
        \vspace{-15pt}
        \caption{Flight time and task success rate of end-to-end fault tolerance analysis by corrupting inter-kernel states.} 
        \label{fig:fault_characterization}
        \vspace{-10pt}
\end{figure}

\vspace{-3pt}
To understand error propagation across kernels, we analyze the impact of corrupted inter-kernel states in the PPC pipeline, which provides insights to improve the PPC kernels and facilitate error detection and mitigation in ~\S\ref{sec:dection_correction}. We do 100 navigation task runs for each evaluation. As shown in Fig.~\ref{fig:fault_characterization}, inter-kernel states exhibit different resilience  and impact on UAV QoF metrics based on their functionality.
For example, in the perception stage, \textit{future\_collision\_seq} is much more robust than \textit{time\_to\_collision}, whose QoF metrics noticeably vary when compared to the golden run. Faults in \textit{time\_to\_collision} can skew the UAV's perceived distance to obstacles. Similarly, data corruption of \textit{(x, y, z)} and \textit{yaw} of way-points planned by motion planner can lead to a wrong direction or crash into obstacles, and faults in \textit{(vx, vy, vz)} could make the UAV fail to keep track of a trajectory. As a result, the distorted trajectory leads to collision or increased flight time and mission energy. 

Bit-flips in different data fields impact UAV behavior differently. Prior works have evaluated data field impact on the processor and neural network~\cite{li2017understanding}, and we further corroborate this in end-to-end UAV systems from the application-level perspective. Our results show that faults in sign and exponent fields have a greater impact on the UAV's resilience and result in increased flight time, energy, and failure cases. We leverage this insight in lightweight UAV anomaly detection in \S\ref{sec:dection_correction}.





\section{Error Detection and Recovery}
\label{sec:dection_correction}
We propose two software-level low-overhead anomaly detection and recovery schemes. The proposed schemes detect anomalous behavior of the inter-kernel states in the PPC pipeline and cease the error propagation, ensuring UAV's safety.

\subsection{Overview of Detection and Recovery}
\label{subsec: overview}

Anomaly detection has been used to distinguish anomaly from normal data distribution in many applications~\cite{ruff2021unifying}. However, there is no effective general anomaly detection technique for different domains. Moreover, autonomous machines are complex systems that typically involve multiple kernels' heterogeneous computing. It is infeasible to separate normal data from anomaly based on the system's input (e.g., sensor readings) and output (e.g., flight commands). The heterogeneity also makes it hard to extract information from the system for anomaly detection. As a consequence, no prior work has focused on anomaly detection to enhance the resilience of UAVs. 

\begin{figure}[t]
\centering
    \subfloat[][Overview. \label{fig:scheme_overview}]{\includegraphics[width=0.48\textwidth]{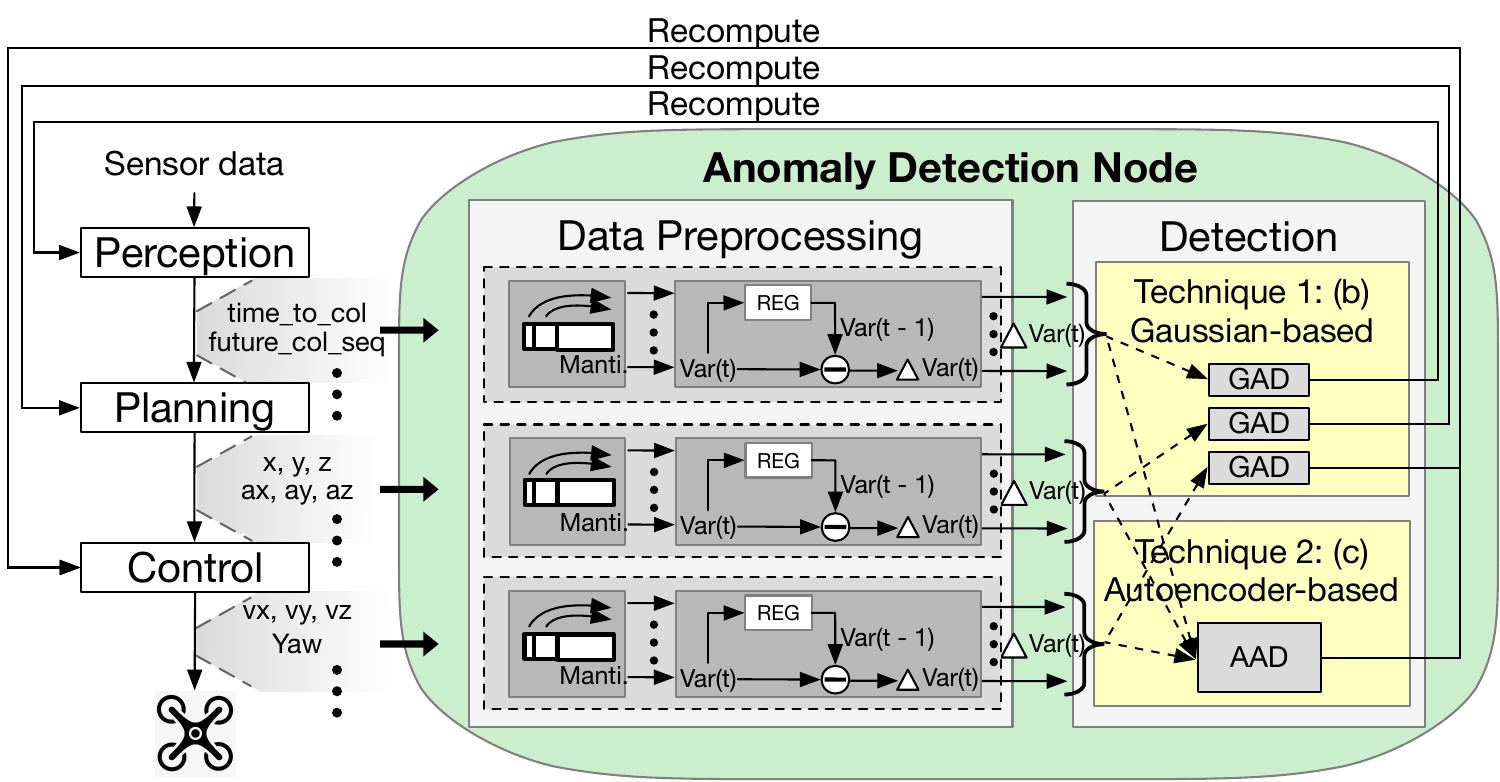}} \\
    \subfloat[][Gaussian-based. \label{fig:Gaussian-based}]{\includegraphics[width=0.202\textwidth]{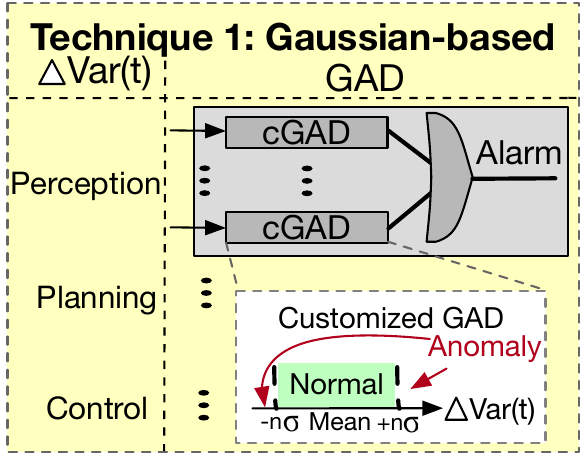}}
    \subfloat[][Autoencoder-based.\label{fig:Autoencoder-based}]{\includegraphics[width=0.298\textwidth]{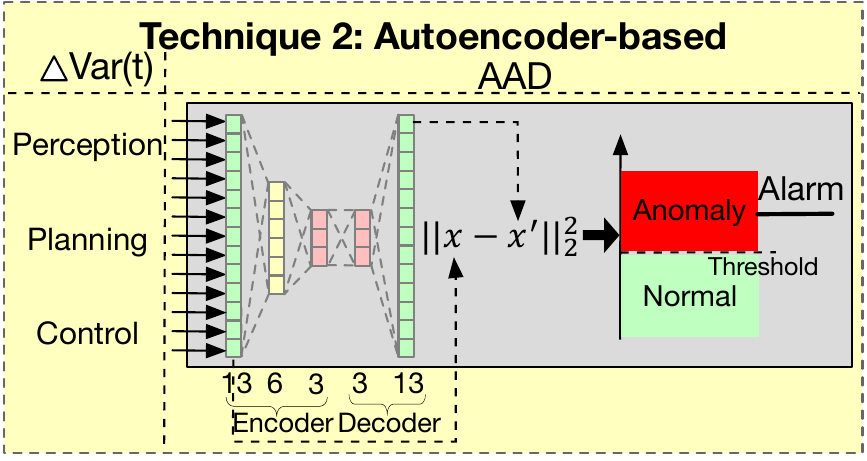}}
    \caption{The proposed anomaly detection and recovery scheme for UAV computational pipeline.}
    \label{fig:detect_correct}
    \vspace{-5pt}
\end{figure}

We propose two anomaly detection techniques to detect SDC that could cause safety hazards for UAVs, including Gaussian- and autoencoder-based techniques. It is observed that both techniques can greatly enhance the safety and resilience of UAVs with low computational overhead.
Fig.~\ref{fig:scheme_overview} shows the proposed anomaly detection and recovery scheme for UAVs. 

According to the analysis in Section~\ref{subsec: cross-stage}, the inter-kernel states, as shown in Fig.~\ref{fig:fault_characterization}, are monitored for anomalous SDC. The monitored states pass their data through a data preprocessing module to increase the detection performance while further reducing the computational overhead. After data preprocessing, the processed states go into either of the proposed anomaly detection techniques for supervision.

Error recovery is a feedback loop from the detection modules to the PPC pipeline. Once an anomalous behavior is detected, an alarm signal will be raised by the detection modules, triggering the recomputation of the corresponding stage, which prevents the corrupted inter-kernel states from propagating to the other kernels. The proposed detection and recovery system can greatly increase the resilience of UAVs' PPC pipeline against SDCs that degrades the safety and flight performance of UAVs. Our approach focuses on SDC as ROS node \textit{crash} can be detected by the ROS system. The ROS master node would restart the node automatically if it crashes. 

\subsection{Data Preprocessing}
\label{subsec: data preprocessing}

In Fig.~\ref{fig:scheme_overview}, the monitored inter-kernel states from the PPC pipeline are processed in the data preprocessing block before sent to the anomaly detection block. Data preprocessing has two steps, including data format transformation and delta calculation. First, for data format transformation, the sign and exponent bits of \textit{float64} states are transformed into 16-bits integer states. Since SDC at the mantissa bits of \textit{float64} is insignificant for value changes, only the sign and exponent bits are monitored to reduce the detection overhead. Second, the deltas of the incoming states are calculated. We define delta as the number of value changes from the previous time point to the current time point for an inter-kernel state. We found that the delta value distribution is close to a Gaussian distribution and has a much smaller value range than the original data, making the differences between normal and anomaly data even larger.


\subsection{Gaussian-based Anomaly Detection}
\label{subsec: gaussian detection}
Fig.~\ref{fig:Gaussian-based} shows the design of the Gaussian-based Anomaly Detection (GAD). Each PPC stage has a corresponding GAD that consists of several customized GAD (cGAD) for each inter-kernel state. If the value of an incoming state is outside the range of its normal data distribution, its cGAD will send out an alarm. The alarms from each cGAD are gathered for each PPC stage, respectively. An alarm from a GAD would trigger the recomputation path of its corresponding stage, stopping the error propagation to the next stage. 

The Gaussian model parameters (i.e., mean, standard deviation) for each cGAD are estimated as following equations:

\vspace{-0.17in}
\begin{equation}
    M_k = M_{k-1} + (x_k - M_{k-1})/k
\end{equation}
\vspace{-0.25in}
\begin{equation}
    S_k = S_{k-1} + (x_k-M_{k-1})(x_k-M_k)
\end{equation}
\vspace{-0.25in}

where $k$ is the number of samples, $M_k$ is the mean value for the $k$ samples, and $S_k$ is an auxiliary term used to compute standard deviation $\sigma$. At initialization, we introduce and set the terms $M_1 = x_1, S_1 = 0$. The parameters are updated online with the recurrence formulas above for new incoming data $x_k$~\cite{knuth2014art}. For $k \geq 2$, the standard deviation $\sigma$ can be derived by $\sigma = \sqrt{S_k/(k-1)}$. Whenever the value of the incoming data is \textit{n} sigma away from the mean value, the alarm of the cGAD will be raised. The number of sigma \textit{n} is a configurable variable that can be optimized based on task complexity.

\subsection{Autoencoder-based Anomaly Detection}
\label{subsec:autoencoder}
Fig.~\ref{fig:Autoencoder-based} shows the Autoencoder-based Anomaly Detection (AAD). The AAD block collects the processed states from all PPC stages as input. An alarm will be raised and triggers the recomputation of the control stage if an anomaly is detected. The proposed autoencoder comprises an encoder with three fully connected (FC) layers of size 13, 6, and 3 neurons, and a decoder with two FC layers of size 13 and 3 neurons. The decoder takes the compressed data from the encoder and outputs the reconstructed input data. The reconstruction error is the difference between the input and output of the autoencoder. We use the mean squared error during the unsupervised training as the reconstruction error minimized by the Adam optimizer. If the reconstruction error is beyond the threshold at the inference phase, the alarm will be raised. The threshold is the upper bound of the reconstruction error in the error-free run.

Rather than a separate Gaussian-based detection module for each PPC stage, we use a single autoencoder for the whole PPC pipeline to leverage the correlation among the inter-kernel states. Once an anomaly is detected, the alarm triggers the recomputation of the control stage. In this way, the autoencoder scheme achieves higher detection performance while reducing the recomputation overhead as shown in~\S\ref{subsec: compute_overhead}.



\subsection{Anomaly Detection and Recovery on ROS}
\label{subsec: AD_on_ROS}
The anomaly detection and recovery scheme is built as a ROS node that contains the data preprocessing and anomaly detection functions. The detection node subscribes to the topics containing the inter-kernel states in the PPC pipeline as input and publishes recomputation signals to the corresponding stages once an anomalous inter-kernel state is detected. The detection node can thus supervise inter-kernel states of the PPC pipeline, avoiding error propagation, thus increasing the resilience of UAV's computational pipeline with negligible overhead.

\section{Experimental Setup}
\label{sec: results}

\begin{figure*}[t!]
\vspace{-0.24in}
  \subfloat[UE Factory.]{
	\begin{minipage}[b][][b]{
	   0.24\linewidth}
	   \includegraphics[width=\textwidth]{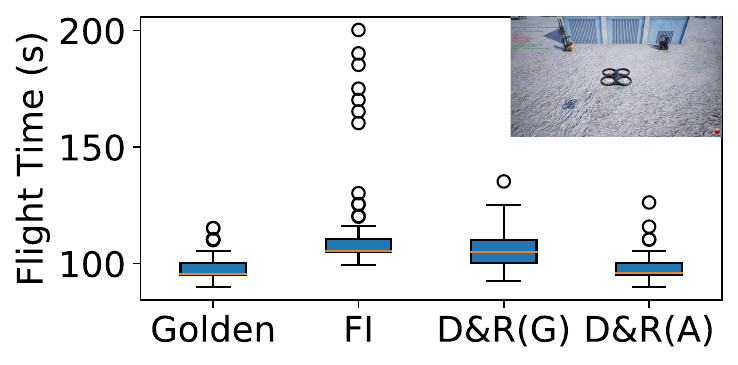}
	   \label{fig:detect_factory}
	   \vspace{-0.2in}
	\end{minipage}}
  \subfloat[UE Farm.]{
	\begin{minipage}[b][][b]{
	   0.24\linewidth}
	   \includegraphics[width=\textwidth]{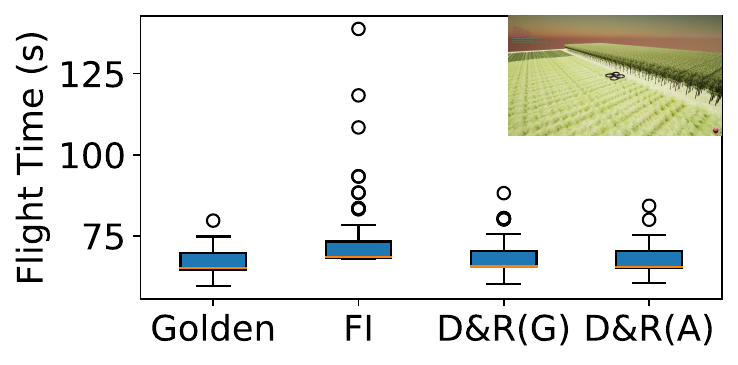}
	   \label{fig:farm}
	   \vspace{-0.2in}
	\end{minipage}}
	  \subfloat[Sparse.]{
	\begin{minipage}[b][][b]{
	   0.24\linewidth}
	   \includegraphics[width=\textwidth]{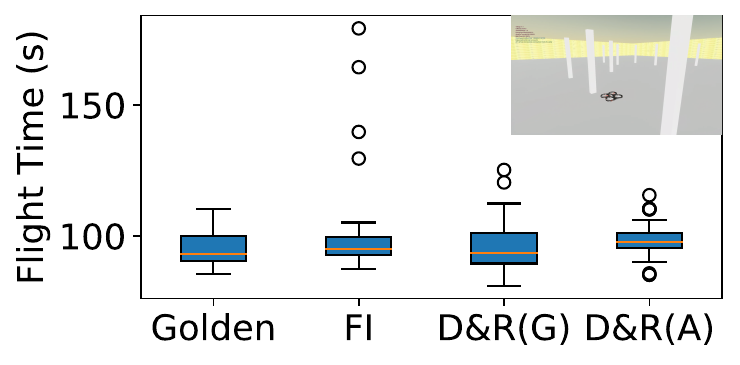}
	   \label{fig:detect_low_dense}
	   \vspace{-0.2in}
	\end{minipage}}
		  \subfloat[Dense.]{
	\begin{minipage}[b][][b]{
	   0.24\linewidth}
	   \includegraphics[width=\textwidth]{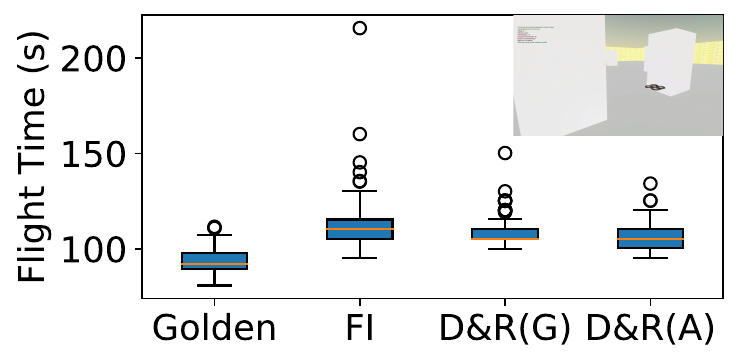}
	   \label{fig:detect_high_dense}
	   \vspace{-0.2in}
	\end{minipage}}
	\vspace{-0.05in}
\caption{Flight time for golden, FI, D\&R(Gaussian), and D\&R(Autoencoder). D\&R is detection \& recovery.}
\label{fig:detection}
\vspace{-10pt}
\end{figure*}

\textbf{Hardware-in-the-loop Simulator.} We use the state-of-the-art closed-loop simulator, MAVBench~\cite{boroujerdian2018mavbench}, as the experimental platform. We use sensors, including RGB-D camera and IMU. An Intel i9 CPU and an NVIDIA 2080 Ti GPU are used as the host machine to simulate environments and the UAV. The companion computer is equipped with an i9 that takes sensory data and generates flight commands. We also evaluated ARM based platforms and the conclusions are unchanged (\S\ref{sec:platforms}). 

\textbf{Evaluation Environments.}
The anomaly detection and recovery schemes are evaluated in four different environments, including \textit{Factory}, \textit{Farm}, \textit{Sparse}, and \textit{Dense}, which are unknown to the UAV. The \textit{Factory} and \textit{Farm} are provided by UE, representing common navigation scenarios with blocks, walls, and hedges. We define [\textit{obstacle density, side length of cuboid obstacles (meters)}] as an environment configuration pair. We generate the \textit{Sparse} with [\textit{0.05, 6}] and the \textit{Dense} with [\textit{0.2, 10}] using a UAV environment generator~\cite{boroujerdian2021roborun}.

\textbf{Training Environments.}  Autoencoder-based and Gaussian-based techniques are trained in a hundred of error-free randomized environments generated by the environment generator. 


\section{Evaluation}
\label{subsec:correct_result}
We run 100 error-free simulations for each environment as the baseline (golden run). Then, we conduct 900 single-bit injections at the instruction level for each environment, including 300 runs for each setting (i.e., \textit{fault injection (FI)}, \textit{detection \& recovery with Gaussian (D\&R(G))}, and \textit{detection \& recovery with autoencoder (D\&R(A))}), as shown in Fig.~\ref{fig:detection}. In each setting, we have 100 fault injections for each PPC stage. Each run includes a one-time single-bit injection. A total of 1000 runs is chosen and each run takes about 5 minutes. The UAV experiment time is a limiting factor for the total runs.

\subsection{Safety Metrics}

\textbf{Improvement of success rate.}
Tab.~\ref{tab:result_recovery} shows the success rates of UAV flights across four environments. In the fault injection runs, the success rate drops 9.7\% in the \textit{Dense} environment. Faults may easily cause collisions or fail to find a collision-free path in complex environments. By contrast, \textit{Farm} is an obstacles-free environment. Even if a UAV detours from its path, there are more feasible paths toward the destination than a complex environment. With the anomaly detection and recovery scheme, Gaussian- and autoencoder-based techniques recover up to 89.6\% and 100\% (fully recover) of failure cases, respectively. Generally, the autoencoder recovers more failure cases than the Gaussian-based scheme and increases the success rate close to or the same as the error-free runs.

\textbf{Improvement of QoF metrics.}
Fig.~\ref{fig:detection} shows the flight time of all successful cases in Tab.~\ref{tab:result_recovery}. 
The fault injection runs result in a much wider range of flight time than the golden run and increase the flight time by 73.8\%, 74.2\%, 62.6\%, and 93.3\% in the worst case for each environment, respectively. However, with Gaussian-based anomaly detection and recovery, many outliers can be recovered, and the worst-case flight time is recovered by 56.4\%, 63.5\%, 49.0\%, and 58.7\%. 
On the other hand, the autoencoder-based technique recovers most of the outliers and can recover the worst-case flight time by 64.2\%, 68.4\%, 57.8\%, and 73.0\%, outperforming the Gaussian method.

\renewcommand{\arraystretch}{1.2}
\begin{table}[b!] \vspace{-0.1in}
\centering
\caption{The flight success rate in 4 evaluation environments.}
\resizebox{0.8\linewidth}{!}{
\begin{tabular}{|l|l|l|l|l|}
\hline
\textbf{Environment} & \textbf{Factory} & \textbf{Farm} & \textbf{Sparse} & \textbf{Dense} \\ \hline
Golden Run           & 100.0\%          & 100.0\%       & 95.0\%          & 85.0\%         \\ \hline
Injection Run        & 91.7\%           & 97.3\%        & 88.3\%          & 75.3\%         \\ \hline
Gaussian-based       & 98.7\%           & 99.3\%        & 94.3\%          & 83.0\%         \\ \hline
Autoencoder-based    & 99.3\%           & 100.0\%       & 95.0\%          & 84.7\%         \\ \hline
\end{tabular}
}
\label{tab:result_recovery}
\end{table}


\textbf{Comparison of Gaussian-based and autoencoder-based schemes.}
The autoencoder-based technique consistently outperforms the Gaussian-based technique in success rate and QoF metrics. The reason is that the autoencoder can leverage the correlation among the inter-kernel states; thus, it can detect the subtle discrepancy of the states. However, the Gaussian-based technique does not have correlation information among states. Therefore, it can only detect each variable separately, which may fail to detect anomalies if the corrupted data is still inside the range of the normal data distribution. 

\textbf{Comparison of environments.} Environments with a higher density of obstacles make it difficult to recover from errors. For the \textit{Dense} environment, a UAV has more complex trajectories to follow and more dynamic actions in response to the obstacles, making the range of the variable distribution wider. The wider distribution increases the number of false-negative detection. Thus, there is still a 20.1\% gap between recovery and golden for the worse case. On the other hand, for the obstacle-free \textit{Farm} or \textit{Sparse} environment, the autoencoder-based technique can achieve a similar performance as the golden run.

\subsection{Flight Trajectory Analysis}
\label{subsec:trajectory}

\begin{figure}[b!] \vspace{-0.15in}
\centering
    \subfloat[][Fault injection in perception.
    \label{fig:traj_time2col}]{\includegraphics[width=0.23\textwidth]{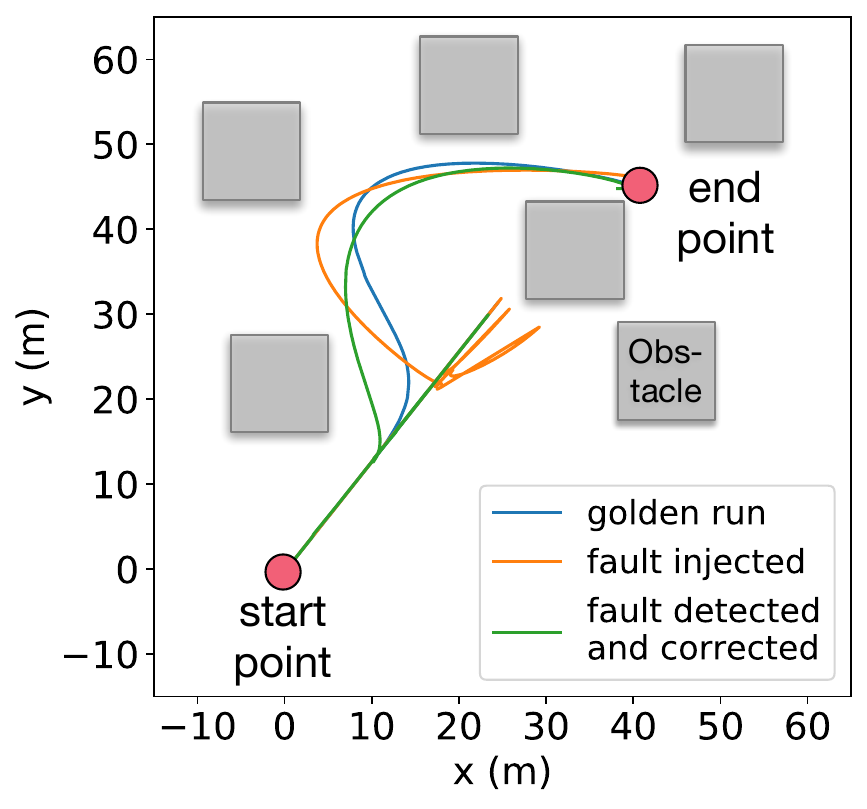}} 
    \hspace{0.05in}
    \subfloat[][Fault injection in planning.\label{fig:traj_xyz}]{\includegraphics[width=0.23\textwidth]{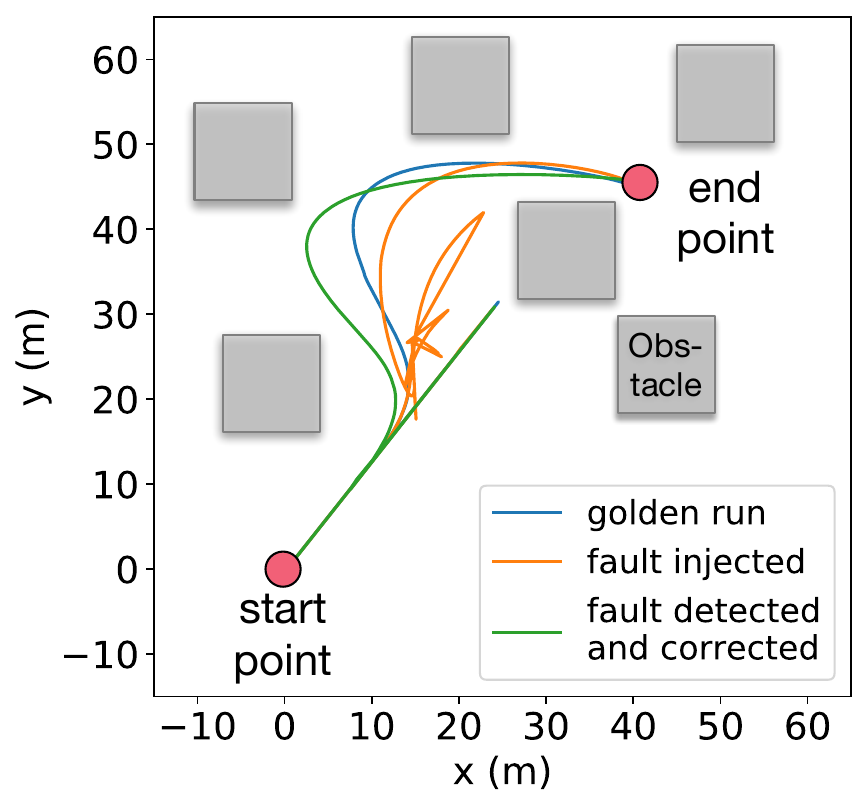}}
    \vspace{-10pt}
    \caption{Trajectories of a golden run, with fault injection, with both fault injection and error detection and recovery.}
    \label{fig:traj}
\end{figure}

To show the impact of faults and the effectiveness of our detection and recovery schemes, we visualize UAV's trajectories in the \textit{Dense} environment. We present the trajectories with the autoencoder-based technique, while the Gaussian-based technique has similar results when successful.

Fig.~\ref{fig:traj} shows the scenario where a single-bit injection in the PPC stage can lead to a flight detour and how the detection and recovery scheme improves the flight. Without fault injection (blue curve), the UAV takes off at the start point and flies towards the endpoint in the beginning phase. Then, when facing an obstacle, it stops at a safe distance and re-plans a new trajectory that flies back slightly and bypasses the obstacle. 

When faults corrupt critical inter-kernel states, such as the coordinate of a way-point, the path may be distorted. The UAV may not stop until it faces an obstacle (orange curve), which causes the UAV to fly back or re-plan its trajectory. The more often the UAV re-plans and detours from its path, the longer it takes to reach the destination, which increases the flight time by 21.9\% and 24.5\% for Fig.~\ref{fig:traj_time2col} and Fig.~\ref{fig:traj_xyz}, respectively. With the detection scheme, the corrupted way-point will be abandoned once an anomaly is detected. The alarm raised by the detection module triggers the stage recomputation. Therefore, the UAV would follow a better trajectory (green curve) without detour.

\begin{table}[t!]
\centering
\caption{Compute time overhead of detection and recovery.}
\renewcommand*{\arraystretch}{1.2}
\resizebox{\linewidth}{!}{%
\begin{tabular}{|c|c|c|c|c|c|c|c|c|}
\hline
\textbf{Environment} & \multicolumn{2}{c|}{\textbf{Factory}}     & \multicolumn{2}{c|}{\textbf{Farm}}        & \multicolumn{2}{c|}{\textbf{Sparse}}      & \multicolumn{2}{c|}{\textbf{Dense}}       \\ \hline
                     & \textbf{DET}                 & \textbf{RECOV}              & \textbf{DET}                  & \textbf{RECOV}               & \textbf{DET}                  & \textbf{RECOV}               & \textbf{DET}                  & \textbf{RECOV}               \\ \hline
\textbf{Perception}           & \textless{}0.0001\% & 0.9603\%           & \textless{}0.0001\% & 1.0902\%           & \textless{}0.0001\% & 0.9788\%           & \textless{}0.0001\% & 1.1932\%            \\ \hline
\textbf{Planning}             & \textless{}0.0001\% & 1.0199\%           & \textless{}0.0001\% & 0.7801\%           & \textless{}0.0001\% & 0.9421\%           & \textless{}0.0001\% & 1.0279\%            \\ \hline
\textbf{Control}              & 0.0008\%           & \textless{}0.0001\% & 0.0007\%           & \textless{}0.0001\% & 0.0009\%           & \textless{}0.0001\% & 0.0012\%           & \textless{}0.0001\% \\ \hline
\textbf{sum (Gaussian)}       & \multicolumn{2}{c|}{1.9810\%}             & \multicolumn{2}{c|}{1.8710\%}             & \multicolumn{2}{c|}{1.9218\%}             & \multicolumn{2}{c|}{2.2223\%}             \\ \hline
\textbf{PPC}                  & 0.0042\%           & \textless{}0.0001\% & 0.0037\%             & \textless{}0.0001\% & 0.0047\%             & \textless{}0.0001\% & 0.0062\%             & \textless{}0.0001\% \\ \hline
\textbf{sum (AutoE)}    & \multicolumn{2}{c|}{0.0042\%}             & \multicolumn{2}{c|}{0.0037\%}             & \multicolumn{2}{c|}{0.0047\%}             & \multicolumn{2}{c|}{0.0062\%}             \\ \hline
\end{tabular}
}
\label{tab:compute_overhead}
\vspace{-10pt}
\end{table}

\subsection{Compute Overhead}
\label{subsec: compute_overhead}
\textbf{Software-level protection.}
Since UAVs can be compute-constrained, we study the overhead of the proposed software-level anomaly detection and recovery scheme across the tested environments. Tab.~\ref{tab:compute_overhead} shows the average overhead of \textit{D\&R(G)} and \textit{D\&R(A)}) in Fig.~\ref{fig:detection}. The autoencoder overhead is much smaller than the Gaussian-based technique's overhead. The overhead of the Gaussian-based technique is dominated by the recovery of perception and planning stages, around 289~\textit{ms} for each occupancy map update and 83~\textit{ms} for each trajectory generation. On the other hand, even if the autoencoder-based technique's detection overhead is higher, the recovery overhead is negligible as the control stage recomputation only takes 0.46~\textit{ms}. As the scheme is operated at the software level with negligible overhead, it is possible to deploy multiple anomaly detection nodes to improve the robustness of detection nodes. 

\begin{figure}[t!]
\centering
    \subfloat{\includegraphics[width=0.48\textwidth]{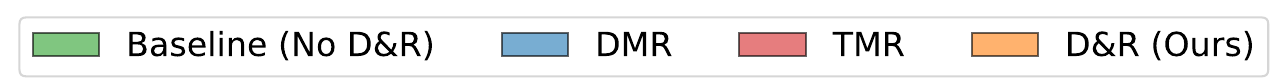}}\\
    \subfloat[][AirSim UAV. \label{fig:DMR_CPS}]{\includegraphics[width=0.24\textwidth]{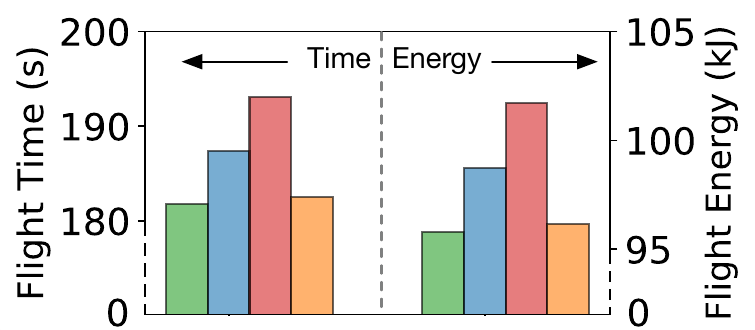}}
    \subfloat[][DJI Spark.\label{fig:DMR_DJI}]{\includegraphics[width=0.24\textwidth]{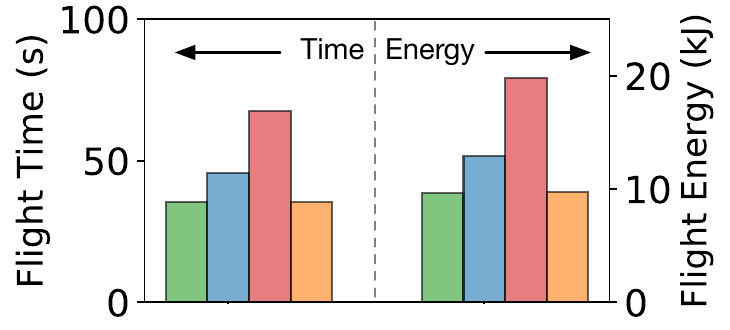}}
    \vspace{-5pt}
    \caption{Comparison of DMR, TMR, and the anomaly detection and recovery schemes on ARM Cortex-A57.}
    \label{fig:DMR_compare}
    \vspace{-15pt}
\end{figure}

\textbf{Hardware-level protection.}
To demonstrate the benefits of our schemes over redundancy-based hardware protections, we adopt a UAV visual performance model from~\cite{krishnan2020sky} to evaluate the performance overhead of microarchitecture-based redundancy schemes (DMR and TMR) on UAV. Two types of UAVs, AirSim UAV and DJI Spark (with the same specs as~\cite{krishnan2020sky}), are used as experimental platforms.
Fig.~\ref{fig:DMR_compare} shows that TMR incurs a flight time increase by 1.06$\times$ on AirSim UAV and 1.91$\times$ on DJI compared to the anomaly detection scheme. The rationale is that hardware redundancy brings higher compute power with higher thermal design power and weight, thus lowering flight velocity and increasing flight time.
Given the tight resource constraints of the UAV system, our scheme demonstrates negligible performance overhead.

\subsection{Computing Platform Comparison}
\label{sec:platforms}
To show the portability we conduct fault injection on different platforms by introducing a single bit-flip at the inter-kernel states. Fig.~\ref{fig:TX2_x86_compare} shows a similar error trend for both platforms. On the TX2, the worst flight time increases 2.8$\times$ since TX2 is an edge platform that has slower responses to environmental changes. However, with the anomaly detection ROS node continuously monitoring the anomaly of inter-kernel states, the flight time is recovered by 79.3\% and 88.0\% with Gaussian-based and autoencoder-based techniques, respectively. 

\newcommand{\txtab}{
\renewcommand{\arraystretch}{1.2}
\resizebox{0.5\linewidth}{!}{
\begin{tabular}{|l|c|c|}
\hline
                     & \textbf{i9-9940X} & \textbf{Cortex-A57} \\ \hline
Core Number         & 14                & 4                   \\ \hline
Core Freq. (GHz) & 3.3               & 2                   \\ \hline
Power (Watt)         & 165               & $<$15                 \\ \hline
Flight time (s)      & 115               & 322                 \\ \hline
Flight energy (kJ)   & 61.7              & 177.1               \\ \hline
\end{tabular}
}
}
\begin{figure}[t!]
    \subfloat{\includegraphics[valign=B, width=0.5\linewidth]{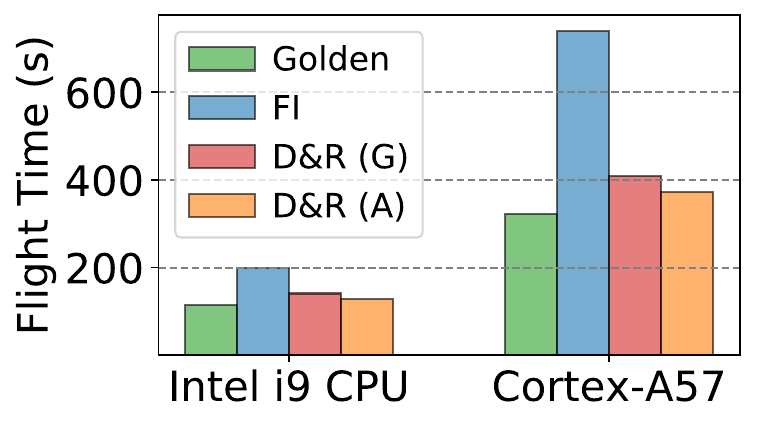}}
    \subfloat{\adjustbox{width=0.5\columnwidth,valign=B,raise=\baselineskip}\txtab}
    \vspace{-5pt}
    \caption{Comparison of detection and recovery schemes.}
    \label{fig:TX2_x86_compare}
    \vspace{-15pt}
\end{figure}

\section{Conclusion}
Safety is paramount for UAVs. Yet, to date, there is no SDC evaluation for them. We present the first fault analysis framework to enable system-level resilience analysis. To enhance the safety and resilience of UAVs, we propose two anomaly detection and recovery schemes and demonstrate that with $<$0.0062\% compute overhead, the autoencoder-based scheme can recover up to 100\% failure cases in the tested scenarios. 



\section*{Acknowledgement}
This work was sponsored in part by the ADA and C-BRIC, two of six centers in JUMP, a Semiconductor Research Corporation (SRC) program sponsored by DARPA.

\pagenumbering{arabic}

\bibliographystyle{IEEEtran}
\bibliography{main}

\end{document}